# The Myth of Culturally Agnostic AI Models

Eva Cetinic
Digital Visual Studies, University of Zurich, Switzerland
eva.cetinic@uzh.ch

AI models trained on enormously large datasets, specifically large language or vision-language models, represent a phenomenon that has by far outgrown the notion of "being just a tool". Regardless of how and for what purpose such models are practically employed, they *in themselves* represent a valuable object of study. "*In themselves*" in this context includes not only all the culturally dependent patterns learned from the training data, but also the culturally dependent attempts to control, modify or erase culturally dependent patterns in the training data. Focusing on the comparative analysis of outputs from two very popular text-to-image synthesis models, DALL·E 2 [1] and Stable Diffusion [2], this paper tries to tackle the pros and cons of striving towards culturally agnostic vs. culturally specific AI models.

Implemented in one way or another, most commonly the "guiding principle" behind many existing text-to-image generators is the ground-breaking vision-language model CLIP and its emerging derivatives (e.g. OpenCLIP [4]). Using enormous amounts of data sampled from the Internet, those models are trained to maximize the similarity between embeddings of existing image-text pairs. Naturally, the models not only learn literal translations between modalities (e.g. the word *cat* and an image of a cat), but also various cultural references (e.g. names of historical monuments or famous people; titles of paintings or movies, etc.), as well as indirectly integrate various (racial, gender, etc.) biases. By encoding numerous associations which exist between data items collected at a certain point in time, those models therefore represent synchronic assemblages of cultural snapshots[1], embedded in a specific technological framework. Metaphorically those models can be considered as some sort of encapsulation of the collective (un)conscious (an occurring recent comparison, e.g. [6, 7]). Text-to-image generators enable us to explore these encapsulations in an interactive and captivating way. In the last two years the development of generative text-to-image models significantly progressed in relation to the quality of generated images, particularly regarding the capacity to depict photorealistic combinations of various objects and concepts. The fact that such models can now render very convincing images triggered a lot of discussion about the potential harm of these technologies and debates about whether they should be openly available for anyone to use. In the context of the "collective unconscious" comparison, one could say that it is yet unclear how to conduct the "shadow work" - in other words, how to confront the repressed, hidden, harmful, violent, problematic aspects of data encoded and propagated through those models.

So how do Stable Diffusion and DALL·E 2 differ in the way they deal with the "dark side" of things? And how does that influence the way they encode and impact culture? Elaborating on the problematic aspects of large language models as "stochastic parrots" that propagate dominant worldviews and biases [8], Underwood outlined the importance of approaching large-scale pretrained models as models of culture [9]. He suggests that while the goal of AI researchers is often directed towards modeling general intelligence, for historians and artists the value of such models lies in their cultural specificity. The potential to use large vision-language and text-to-image models for empirical cultural studies relies on the understanding of the relation between (cultural) generality and specificity encoded within the models. However, the goal of limiting the potentially undesirable capabilities of such models results in attempts

---

[1] "Cultural snapshots are recorded samples of public environments commonly encountered by many people" [5]

to build culturally agnostic models. But is it possible to create truly culturally agnostic models, and should that be the goal? A preliminary way to explore this question is to analyze, compare and interpret outputs of different text-to-image models.

The comparison of Stable Diffusion and DALL·E 2, indicates that there is a trade-off between risk mitigation and cultural specificity. The developers of the Stable Diffusion model take a more "libertarian" approach - being the first to make their models open access without restrictions. Also, the LAION dataset [10] on which the models have been trained is openly available for exploration. Several versions of Stable Diffusion models are introduced, and some of them have been fine-tuned on filtered versions of the LAION dataset (filtered based on the estimated aesthetic score). However, the base model is created using an unfiltered version of the dataset and is made available with the following warning remark: "*Stable Diffusion v1 is a general text-to-image diffusion model and therefore mirrors biases and (mis-)conceptions that are present in its training data.*" [11] On the other hand, the by OpenAI developed model DALL·E 2 was introduced with restricted accessibility (e.g. waiting list) and developed with the aim to reduce potential risk associated with the model. DALL·E 2 was trained on a filtered version of their (unpublished) large image-text pair dataset and specific pre-training mitigation techniques [12] were introduced in order to (1) remove violent and sexual content; (2) remove the consequent filter-induced (e.g. gender related) biases; (3) avoid memorization. In relation to cultural specificity, the most interesting aspect is their aim to tackle the issue of memorization, in particular the problem of *image regurgitation* which is caused by images that are replicated many times in the dataset. Removing images that appear many times in the dataset basically means dissolving many culturally dependent associations between words and images. For example, the title "Starry night" is in the western-art tradition immediately associated with Van Gogh's famous painting. Because it is famous and well-known, the image of the painting is not only included in many websites, but has been reproduced numerous times on T-shirts, posters, stickers, home decor, etc. Surely near duplicate images of all these products can be found in the unfiltered datasets, paired with the text "Starry night", which strengthens this association of specific image features and words in the latent space. Figure 1 shows some examples of images that show how specific culturally dependent associations between images and titles/names are preserved in Stable Diffusion, but not in DALL·E 2.

**Figure 1** Images generated using DALL·E 2 (top row) and Stable Diffusion (bottom row) generated from the following prompts (number corresponds to the column): 1) *Starry night*; 2) *Portrait of Carl Gustav Jung sailing on the Zurich lake, historical photo, black and white, photorealistic*; 3) *Dark side of the moon*; 4) *The garden of earthly delights*; 5) *The cure*; 6) *Magical mystery tour*.

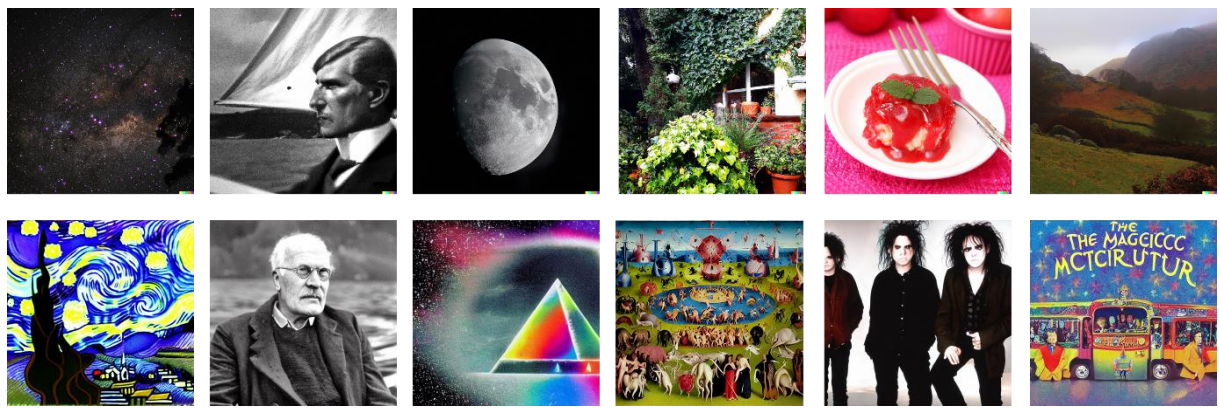

If the goal is to generate neutral outputs and avoid that the training data *regurgitates* iconic and well-known image-text pairs, then it makes sense to implement techniques that eliminate data memorization. However, when such memorizations are preserved and especially when they emerge in unexpected or polysemic combinations, then not only is it more interesting to interact with such a model, but such a model also holds much more potential for empirical cultural studies based on synthetic data. One of the reasons why the community of AI artists and text-to-image enthusiasts found the Stable Diffusion model particularly interesting is because the notion of memorization is more preserved and therefore the model has the ability to reproduce specific objects, scenes or faces of famous people. The DALL·E 2 model is in that sense less culturally specific and produces visual outputs where the relation between image and text is more literal. Although DALL·E 2 is less culturally specific, it is still not completely culturally-agnostic. Not only does DALL·E 2 also integrate some specific cultural memories (e.g. being able to reproduce images in the style of some well-known artists), but it also expands cultural dependency from the training data to the choice of filtering the training data. The way DALL·E 2 pre-training mitigations were performed, the choices that were made in deciding what is appropriate and what violates the defined content policies, is also grounded in the specific contemporary cultural and political orientation of western (most commonly Silicon Valley-based) companies and research institutes. Even if this orientation is striving to be ethical and inclusive, it is nevertheless culturally dependent. Taking a position of how to address bias cannot be a neutral position, it is always conditioned by our cultural and historical understanding of ethical principles. And sometimes the noble aim to eliminate bias just inverts the hierarchies within the same type of bias. Such biases are not often easily detectable, but Figure 2 shows one such particular example where Stable Diffusion generates mostly "western-centric" outputs (mostly related to Christian iconography), while DALL·E 2 (probably with the aim to eliminate the west-centric bias), generates mostly "eastern-centric" outputs on a prompt related to religion.

**Figure 2.** Examples of images generated from the prompt "*Songs of faith and devotion*" using DALL·E 2 (top row) and Stable Diffusion (bottom row)

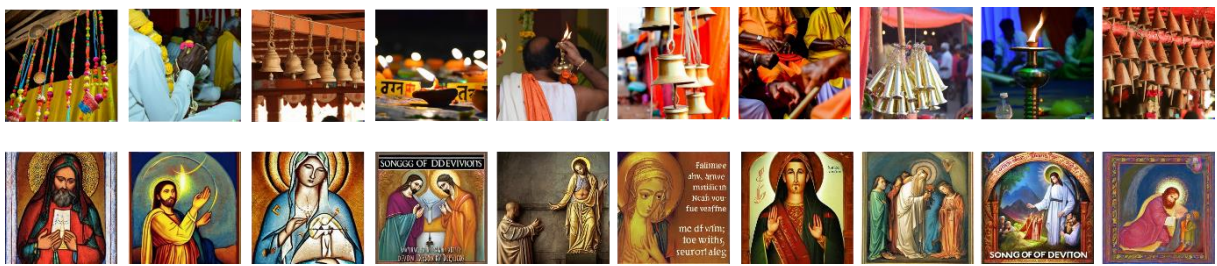

The examples shown in Figure 2 indicate how extremely complex it is to mitigate biases and develop not necessarily culturally agnostic, but culturally inclusive models. However, if we do not consider the applicative aspect of those models (e.g. in what kind of apps are they going to be integrated), then perhaps we should deconstruct the myth of aiming towards unbiased and culturally-agnostic models, and instead focus on how such models can help us understand our cultural specificity, with all its darks sides and shadows. And in this context, the most interesting aspects of text-to-image models relate to the depiction of culturally dependent concepts. A recent analysis on the "*The Datafication of a Kiss*" [7] provides an example of possible approaches. There are numerous other interesting concepts that deserve similar attention. But to systematically study those models as objects of culture we need new methodologies and novel, integrative and interdisciplinary approaches. How to use and understand something that simultaneously resembles a magnifying glass, a mirror and a kaleidoscope? The quest to systematically study culturally dependent encodings in the latent space emerges as the new great research challenge.